\documentclass[graybox]{svmult}

\usepackage{type1cm}        
\usepackage{makeidx}         
\usepackage{graphicx}        
\usepackage{multicol}        
\usepackage[bottom]{footmisc}

\usepackage{newtxtext}       %
\usepackage[varvw]{newtxmath}       

\makeindex             

\begin{document}

\title*{Probabilistic Linguistic Knowledge and Token-level Text Augmentation}
\author{Zhengxiang Wang}
\institute{Zhengxiang Wang \at Department of Linguistics, Stony Brook University, Stony Brook, NY, United States, \\ \email{zhengxiang.wang@stonybrook.edu}}

%
%
\maketitle

\abstract*{This paper investigates the effectiveness of token-level text augmentation and the role of probabilistic linguistic knowledge within a linguistically-motivated evaluation context. Two text augmentation programs, REDA and REDA$_{NG}$, were developed, both implementing five token-level text editing operations: Synonym Replacement (SR), Random Swap (RS), Random Insertion (RI), Random Deletion (RD), and Random Mix (RM). REDA$_{NG}$ leverages pretrained $n$-gram language models to select the most likely augmented texts from REDA's output. Comprehensive and fine-grained experiments were conducted on a binary question matching classification task in both Chinese and English. The results strongly refute the general effectiveness of the five token-level text augmentation techniques under investigation, whether applied together or separately, and irrespective of various common classification model types used, including transformers. Furthermore, the role of probabilistic linguistic knowledge is found to be minimal.}

\abstract{This paper investigates the effectiveness of token-level text augmentation and the role of probabilistic linguistic knowledge within a linguistically-motivated evaluation context. Two text augmentation programs, REDA and REDA$_{NG}$, were developed, both implementing five token-level text editing operations: Synonym Replacement (SR), Random Swap (RS), Random Insertion (RI), Random Deletion (RD), and Random Mix (RM). REDA$_{NG}$ leverages pretrained $n$-gram language models to select the most likely augmented texts from REDA's output. Comprehensive and fine-grained experiments were conducted on a binary question matching classification task in both Chinese and English. The results strongly refute the general effectiveness of the five token-level text augmentation techniques under investigation, whether applied together or separately, and irrespective of various common classification model types used, including transformers. Furthermore, the role of probabilistic linguistic knowledge is found to be minimal.}

\section{Introduction}

Data serves as a crucial component in training high-performing and robust machine learning models that can effectively tackle real-world learning tasks. However, data availability is often unpredictable and not guaranteed. In the realm of supervised learning, the development of reliably deployable models typically requires the collection of vast amounts of annotated data, which is affordable only for a select few. In low-resource settings, in particular, the available data may be limited or entirely nonexistent. There are also situations where existing data is imbalanced for specific classes, causing models trained on such data to be easily biased towards classes with abundant training examples. This can potentially be harmful when the models are deployed. Practical considerations like these have given rise to data augmentation, a widely adopted strategy to mitigate the problems of scarce or imbalanced data. Data augmentation involves applying label-preserving transformations to existing data to generate novel labeled data. This approach has seen considerable success in various fields, such as image and speech recognition \cite{Simard2003, alexnet2012, ko15_interspeech, Cui2015,Park2019,Shorten2019ASO, Iwana2021}.

Text augmentation, a subcategory of data augmentation that focuses on augmenting text data, is a promising yet challenging domain within NLP \cite{Shorten2021, feng-etal-2021-survey,Liu:9240734, yang-etal-2022-learning, sahin-2022-augment, empirical_survey}. The challenge arises due to the lack of well-established methods that can consistently generate diverse and accurate augmented texts simultaneously. In contrast to images or speech, where physical features can be relatively easily manipulated without altering the label, text is highly sensitive to the arrangement and combination of words. For instance, to augment an image, one can rotate, crop, flip, or change its color specifications in a predetermined manner, while still assuming that the augmented images represent the same object as the original \cite{Simard2003}. However, when augmenting text, one cannot merely replace and shuffle words in an automated fashion to generate paraphrases. It becomes evident that there is a need for foundational research exploring the factors that influence the effectiveness of text augmentation \cite{feng-etal-2021-survey}.

The primary objective of this study is to gain a deeper understanding of the effectiveness of token-level text augmentation within a linguistically-motivated evaluation context. Token-level text augmentation is assessed, as opposed to other more complex methods (see Sec~\ref{sec:related_works}), due to its applicability across various tasks and languages. The insights regarding its effectiveness may prove valuable in low-resource domains, where text augmentation is primarily employed in real-world scenarios. More specifically, this study aims to address the following two research questions: (1) How effective is token-level text augmentation? (2) Can the incorporation of probabilistic linguistic knowledge enhance the effectiveness of text augmentation?

To address these two research questions, comprehensive experiments were conducted on a binary question matching classification task, involving both Chinese and English languages, at a reasonably large scale. The objective of this task is to predict whether two given questions share the same expressed intent. This type of task, which entails the classification of text pairs, is well-suited for evaluating text augmentation as it demands high fidelity of the augmented texts to the original texts to maintain label preservation. Conversely, since token-level text augmentation is not strictly paraphrastic, its success in such tasks serves as strong evidence of its overall effectiveness. To explore the impact of probabilistic linguistic knowledge, pretrained $n$-gram language models are also utilized to select augmented texts, which, in theory, should be statistically more likely and expected to be closer to natural texts or of higher quality. Consequently, it is anticipated that probabilistic linguistic knowledge will enhance the effectiveness of text augmentation.


The paper proceeds as follows. Sec~\ref{sec:related_works} reviews related works, while the augmentation methods and experimental settings of the study are detailed in Sec~\ref{sec:aug_methods} and Sec~\ref{sec:exp_settings}, respectively. Sec~\ref{sec:main_results} presents the results of the main experiments, and the findings of three supplementary experiments are reported in Sec~\ref{sec:sup_exps}. Sec~\ref{sec:discussion_conclusion} offers further discussion on the discoveries and provides a conclusion.


\section{Related Works\label{sec:related_works}}

Over the years, three major types of text augmentation have been employed in NLP to generate label-preserving data \cite{empirical_survey}: token-level augmentation, sentence-level augmentation, and hidden-level augmentation. Token-level augmentation focuses on individual tokens and involves word replacements, which can be either dictionary-based \cite{Zhang2015} or embedding-based \cite{wang-yang-2015-thats}, as well as deletion, insertion, or shuffling of tokens in a random \cite{wei-zou-2019-eda} or predefined \cite{kang-etal-2018-adventure, asai-hajishirzi-2020-logic} manner. Sentence-level augmentation, on the other hand, typically entails paraphrasing at the sentence level. Back translation \cite{sennrich-etal-2016-improving, edunov-etal-2018-understanding, Singh2019}, a widely popular technique that involves translating a text into another language and then retranslating it back, exemplifies this approach. Additionally, researchers have utilized language models to generate novel data conditioned on given text or labels \cite{hou-etal-2018-sequence, kobayashi-2018-contextual, kurata16b_interspeech}. Lastly, hidden-level augmentation pertains to the manipulation of hidden representations, such as linear transformations, in order to create new perturbed representations \cite{chen-etal-2020-mixtext, puzzleMix2020, chen-etal-2022-doublemix}.

Many of the aforementioned studies have reported slight yet inconsistent performance gains when training models with augmented data for their respective NLP tasks, mostly text classification tasks. A common explanation for any observed performance improvement is that the augmented data introduces noise to the original training data, thus preventing the trained models from overfitting \cite{xie2020unsupervised}. This, in turn, improves their performance on test sets. 

A notable and widely cited example is provided by \cite{wei-zou-2019-eda}. The paper employs four simple token-level text editing operations to augment train sets of varying sizes and demonstrates their general effectiveness in boosting model performance across five sentiment-related and text type classification tasks. Although claimed to be universal, these four text editing operations, which are also examined in this study (see Sec~\ref{sec:aug_methods}), have been found not to be consistently beneficial. More specifically, they have been shown to negatively impact model performance in more complex tasks, such as natural language inference and paraphrasing \cite{empirical_survey}, and fail to consistently improve performance for transformers \cite{longpre-etal-2020-effective}. 

This study aims to enrich the existing literature on the effectiveness of token-level text augmentation by conducting comprehensive and fine-grained cross-linguistic experiments for an under-explored task. It additionally examines the role of probabilistic linguistic knowledge, also an under-explored yet fundamental question.

\section{Augmentation methods\label{sec:aug_methods}}

\subsection{Text augmentation techniques}

In this study, five token-level text augmentation techniques, or text editing operations, are employed: Synonym Replacement (SR), Random Swap (RS), Random Insertion (RI), Random Deletion (RD), and Random Mix (RM).

The first four techniques were initially proposed by \cite{wei-zou-2019-eda} as a simple but universal set of text augmentation techniques named as EDA (Easy Data Augmentation). For a single text edit, they work as follows. SR randomly replaces a word, where possible, with one of its randomly sampled synonyms (if more than one) based on a predefined dictionary. RS, on the other hand, swaps the positions of a random word pair within a text. RI inserts a random synonym immediately after an eligible target word, while RD deletes a word at random. For $n$ text edits, these techniques are simply applied $n$ times. Additionally, RM, introduced in \cite{wang-2022-linguistic}, is a random combination of 2-4 of the other four techniques, resulting in a more diversified text. Given their randomness, these techniques are also referred to as random text perturbations \cite{wang-2022-random}.

The five text augmentation techniques are implemented in Python using a program called REDA (Revised EDA). In addition to incorporating an extra text editing operation (RM), REDA differs from EDA in three key aspects. Firstly, REDA prevents duplicates in the output text(s), which can occur when there are no synonyms available for replacement (SR) or insertion (RS) for words in the input text, or when the same words are replaced or swapped back during the SR and RS operations. Secondly, REDA does not preprocess the input text (e.g., removing stop words), as this is believed to have minimal impact and better aligns with the fundamental concept of random text perturbations that underlie these augmentation techniques. Lastly, REDA replaces only one word with its synonym at a given position per text edit, rather than replacing all occurrences, which are regarded as additional edits.

In this study, the synonym dictionary for English is derived from WordNet \cite{wordnet}, while for Chinese, it is obtained from multiple reputable sources through web scraping\footnote{\url{https://github.com/jaaack-wang/Chinese-Synonyms}}. Furthermore, rather than applying a fixed number of text edits to texts of varying lengths, this study employs an \emph{editing rate}, which adjusts the number of text edits proportionally to the text length.


\subsection{$N$-gram language model}

A $n$-gram is a string of $n$ words $W_{1}^{n} = (w_1, ... w_n)$. A $n$-gram language model is a Markov model that estimates the probability (typically expressed in logarithmic terms \cite{Jurafsky2009}) of a given string $W_{1}^{L}$ of length $L$ (where $L \geq n$) by the product of the probabilities of all its $n$-long substrings. 

\begin{equation} \label{eq:ngram_lm}
    \log P(W_{i}^{L}) \approx \log \prod_{i=1}^{L-n+1} P(W_{i}^{i+n-1}) = \sum_{i=1}^{L-n+1} \log P(W_{i}^{i+n-1})
\end{equation}

\noindent where $P(W_{i}^{i+n-1})$ represents $P(w_{i+n-1}|W_{i}^{i+n-2})$. Let $C(W_{i}^{i+n-1})$ denote the frequency of occurrence of a string $W_{i}^{i+n-1}$ in a much larger corpus used as the training data. The maximum-likelihood probability estimate for $P(W_{i}^{i+n-1})$ is the relative frequency of $W_{i}^{i+n-1}$ against its previous $(n-1)$ words $W_{i}^{i+n-2}$ in counts:

\begin{equation}
    P(W_{i}^{i+n-1}) \approx \frac{C(W_{i}^{i+n-1})}{C(W_{i}^{i+n-2})}
\end{equation}

\noindent Since both $C(W_{i}^{i+n-1})$ and $C(W_{i}^{i+n-2})$ can be 0 during the deployment of a pretrained $n$-gram language model, this leads to inaccurate or undefined probability estimates. Inspired by both Eq.~(\ref{eq:ngram_lm}) and Stupid Backoff \cite{brants-etal-2007-large}, this study uses a non-discounting method that estimates the probability of an unseen $n$-gram by multiplying the probabilities of its two $(n-1)$-grams together, as shown in Eq.~(\ref{eq:backoff}). The method will continue to back off into unigrams if all other higher-order $n$-grams do not occur in the training data, where unseen unigrams are simply assigned the same probability as those one-off unigrams.

\begin{equation} \label{eq:backoff}
P(W_{i}^{i+n-1}) \approx 
\left\{
    \begin{array}{lc}
        \frac{C(W_{i}^{i+n-1})}{C(W_{i}^{i+n-2})}, & \text{if } C(W_{i}^{i+n-1}) > 0 \\ 
        \prod_{i=1}^{2}P(W_{i}^{i+n-2}), & \text{otherwise }
    \end{array}
\right\}
\end{equation}

In this study, I trained both the Chinese and English $n$-gram language models using $n$-grams up to 4-grams. The pretrained $n$-gram language models are utilized as a filter to select the $k$ most likely outputs from $m$ possible outputs generated by REDA, where $m$ is at least 20 times greater than $k$ in this study. The program that combines REDA with a $n$-gram language model is denoted as REDA$_{NG}$.

\section{Experimental settings\label{sec:exp_settings}}

\subsection{\label{sec:task_and_data}Task and data}

The task under analysis is a binary classification task on the basis of text pairs, commonly known as question matching. The aim is to predict whether a given question pair $(Q, Q')$ express similar intents, which is a fundamental sub-task for question answering or, more broadly, a downstream task for semantic matching \cite{liu-etal-2018-lcqmc}. Two labels are used, with 0 denoting negative match of $(Q, Q')$ and 1 positive match. 

This study considers two large-scale benchmark datasets for question matching. One is the Large-scale Chinese Question Matching Corpus (LCQMC, \cite{liu-etal-2018-lcqmc}) for Chinese. The other is the Quora Question Pairs Dataset (QQQD)\footnote{\url{https://quoradata.quora.com/First-Quora-Dataset-Release-Question-Pairs}} for English. 

For LCQMC, I reused the original train, development, and test sets as provided by \cite{liu-etal-2018-lcqmc}. For QQQD, three label-balanced data sets were created from its train set since the test set is made unlabeled for online competition. The basic statistics about these two datasets are given in Table~\ref{tab:data}.

\begin{table}
\centering
\caption{\label{tab:data}
Statistics of the data splits for LCQMC and QQQD.}
\begin{tabular}{p{1.5cm}p{3.5cm}p{3.05cm}}
\noalign{\smallskip}\hline\noalign{\smallskip}

\textbf{Split} & \textbf{LCQMC} & \textbf{QQQD} \\
 &  (Matched \& Mismatched) & (Matched \& Mismatched) \\ \noalign{\smallskip}\hline\noalign{\smallskip}
 
Train & 238,766 & 260,000 \\
 &  (138,574 \& 100,192)  &  (130,000 \& 130,000) \\ \noalign{\smallskip}\hline\noalign{\smallskip}
Dev & 8,802 & 20,000 \\
 &   (4,402 \& 4,400) &  (10,000 \& 10,000)  \\ \noalign{\smallskip}\hline\noalign{\smallskip}
Test & 12,500 & 18,526 \\

 &  (6,250 \& 6,250) &  (9,263 \& 9,263)  \\ \noalign{\smallskip}\hline\noalign{\smallskip}
 
 \end{tabular}
\end{table}

\subsection{Classification models}

In the main experiments, four classic neural network models were chosen: Continuous Bag of Words (CBOW, \cite{cbow}), Convolutional Neural Network (CNN, \cite{kim-2014-convolutional}), Gated Recurrent Units (GRU, \cite{cho-etal-2014-learning}) and Long Short-Term Memory (LSTM, \cite{lstm}). Since the focus here is to evaluate the effectiveness of token-level text augmentation and the role of probabilistic linguistic knowledge, the use of various classification models is not meant to contrast the learning difference among them, but rather to make the examination more comprehensive. Pretrained word embeddings were not utilized to simulate low-resource settings. For the same reason, transformers \cite{tranformer} were only used in the supplementary experiments, instead of the main ones.      

The models can also be divided into three groups, depending on the type of train sets they train on. \emph{Baseline models} refer to models that train on train sets without augmentation, i.e., train sets that only contain original training examples. Models that train on train sets augmented by REDA are called as \emph{REDA models}, and similarly \emph{REDA$_{NG}$ models} are the ones that train on train sets augmented by REDA$_{NG}$. REDA models and REDA$_{NG}$ models are also called \emph{augmented models}, since both are trained on augmented train sets. Augmented train sets contain augmented examples on the top of the original examples, based on which the augmented examples are produced. For convenience, the three respective types of train sets are simply called \emph{baseline train sets}, \emph{REDA train sets}, and \emph{REDA$_{NG}$ train sets}.

\subsection{Training details}

The models were constructed in PaddlePaddle\footnote{\url{https://www.paddlepaddle.org.cn/en}}, a deep learning framework developed by Baidu and trained on Baidu Machine Learning CodeLab's AI Studio with Tesla V100 GPU and 32G RAM. The models were trained using mini batches of size 64 with the objective of reducing cross-entropy loss. The Adam optimizer \cite{adam} with 5e-4 learning rate was applied to assist training. The training time was consistently 3 epochs long, since most of the models overfitted the train sets within 3 epochs. Development sets were used for validation purposes.

The basic structure for the classification models is simple and unified as follows. Each model begins with two separate embedding layers with the same embedding size to convert the input text pair $(Q, Q')$ into two respective embedded sequences, $\textbf{Embd}_Q$ and $\textbf{Embd}_{Q'}$. Then, $\textbf{Embd}_Q$ and $\textbf{Embd}_{Q'}$ each pass through an encoder layer, whose structure is determined by the classification model in use, to obtain two encoded sequences, $\textbf{Enc}_Q$ and $\textbf{Enc}_{Q'}$. The encoded sequences, $\textbf{Enc}_Q$ and $\textbf{Enc}_{Q'}$, are concatenated along the last axis and then passed to a fully connected feed-forward network (FFN) that consists of two linear transformations with a tanh activation function in between. 

\begin{equation}
FFN(\textbf{x}) = tanh(\textbf{x} \textbf{W}_1 + b_1) \textbf{W}_2 + b_2
\end{equation}

\noindent For CBOW, the encoder is the point-wise summation of the embeddings of tokens in the input text pair followed a tanh function. For the rest, the encoder layers are simply CNN layer, GRU layer, and LSTM layer, corresponding to the model names.

\subsection{Augmentation details}

Due to experimental costs, it was not possible for this study to evaluate the effects of different initializations of REDA/REDA$_{NG}$ (i.e., editing rate, number of augmentation per example) on the trained models’ performance. Therefore, I initialized REDA/REDA$_{NG}$ with small editing rates, informed by \cite{wei-zou-2019-eda}, who recommend small editing rates over large ones and demonstrate that large editing rates lead to performance decline in their ablation experiments. This makes sense since large editing rates are more likely to cause label changes. Intuitively, if small editing rates do not work well, larger ones will not either. The number of augmentation per example, to be mentioned below, was kept small for the same consideration.

More concretely, REDA and REDA$_{NG}$ were initialized with the following editing rates for SR, RS, RI, and RD, respectively: 0.2, 0.2, 0.1, and 0.1. I applied Python rounding rule to calculate and perform the number of edits needed for each operation. That means, if the number of edits is less than or equal to 0.5, it will be rounded down to 0 and thus no editing operation will apply. To make the experiments more controlled and doable, (1) I made RM only randomly perform two of the other four editing operations with one edit each; (2) and every editing operation produced up to 2 non-duplicated augmented texts per text (or 4 per text pair), if the train set size was less than 50k; otherwise, there would only be one augmented text per text instead. Every augmented text was crossed-paired with the other text that was the pair to the text being augmented, with the original label retained for the augmented text pair. These settings were also applied for the supplementary experiments. 

Table~\ref{tab:aug_data} shows the size of the augmented train sets for the main experiments. 

\begin{table}[h]
\centering
\caption{\label{tab:aug_data}
Size of augmented train sets for the main experiments on LCQMC and QQQD. For convenience, 240k is hereafter used to refer to the full size (i.e., 238,766 to be exact) of LCQMC. Note that, all the subsets of the full train sets were randomly and independently sampled.}
\begin{tabular}{p{2cm}p{2cm}p{1cm}p{2cm}p{1.5cm}}
\noalign{\smallskip}\hline\noalign{\smallskip}
\textbf{LCQMC} & \textbf{Augmented} & &\textbf{QQQD} & \textbf{Augmented} \\ \noalign{\smallskip}\hline\noalign{\smallskip}
5k & 66,267 & & 10k & 148,341 \\ \noalign{\smallskip}
10k & 132,513 & & 50k & 543,066 \\ \noalign{\smallskip}
50k & 563,228 & & 100k & 1,086,063 \\ \noalign{\smallskip}
100k & 929,176 & & 150k & 1,629,178 \\ \noalign{\smallskip}
240k & 2,218,512 & & 260k & 2,823,733 \\ \noalign{\smallskip}\hline\noalign{\smallskip}

\end{tabular}
\end{table}

\section{Main experiments\label{sec:main_results}}

This section reports the test set performance of the four classification models trained on train sets of varying size with and without augmentation for the binary question matching task in Chinese and in English. As the test sets for LCQMC and QQQD are equally balanced across labels (see Sec~\ref{sec:task_and_data}), accuracy is considered as the primary evaluation metric. The average precision and recall are taken as secondary metrics for more nuanced analyses.

 \subsection{Chinese: LCQMC}

Table~\ref{tab:accu} shows the test set accuracy of the four classification models trained on the three types of train sets (baseline, REDA, and REDA$_{NG}$) of different sizes. Contrary to the expectation, incorporating probabilistic linguistic knowledge into the five token-level text augmentation techniques does not lead to superior model performance, as the REDA$_{NG}$ models never outperform the REDA models in terms of average performance, when given the same amounts of training data. Instead, the REDA models almost always achieve slightly better performance than the REDA$_{NG}$ counterparts. Moreover, it appears that the augmented models (REDA and REDA$_{NG}$) do not necessarily have better test set accuracy than that of the baseline models, unless augmentation is applied to sufficient original training examples (i.e., at least 50k). 

\begin{table*}[h]
\centering
\caption{\label{tab:accu}
Test set accuracy (\%) of the four classification models trained on LCQMC's train sets of varying size with and without augmentation. The header denotes train set sizes in terms of original training examples. Best performance given a train set size and a model type is highlighted in \textbf{bold}.}

\begin{tabular}{p{2.5cm}p{1cm}p{1cm}p{1cm}p{1cm}p{0.6cm}}
\noalign{\smallskip}\hline\noalign{\smallskip}

Models & 5k & 10k & 50k & 100k & 240k \\ \noalign{\smallskip}\hline\noalign{\smallskip}

CBOW & \textbf{59.4}  & 60.4  & 65.4  & 67.8  & 73.8   \\ 
+REDA & 58.1  & \textbf{60.9}  & \textbf{68.2}  & \textbf{72.2}  & \textbf{76.4}  \\ 
+REDA$_{NG}$ & 58.8  & 59.6  & 68.1  & 71.2  & 76.0 \\ \noalign{\smallskip}\hline\noalign{\smallskip}

CNN & 59.3  & \textbf{63.4}  & 67.2  & 69.0  & 72.9  \\ 
+REDA & 59.8  & 62.6  & 66.8  & \textbf{69.8}  & \textbf{74.9}  \\ 
+REDA$_{NG}$ & \textbf{60.3}  & 62.0  & \textbf{67.9}  & 69.1  & 74.0 \\ \noalign{\smallskip}\hline\noalign{\smallskip}

LSTM & \textbf{60.0}  & \textbf{62.1}  & 66.2  & 69.6  & 74.8  \\ 
+REDA & 58.9  & 61.5  & \textbf{67.7}  & \textbf{71.8}  & \textbf{76.4} \\ 
+REDA$_{NG}$ & 57.7  & 60.9  & \textbf{67.7}  & 71.7  & 75.9 \\ \noalign{\smallskip}\hline\noalign{\smallskip}

GRU & \textbf{59.8}  & \textbf{61.9}  & 68.1  & 70.3  & \textbf{76.8}  \\ 
+REDA & 58.7  & 61.3  & \textbf{68.7}  & \textbf{72.7}  & \textbf{76.8} \\ 
+REDA$_{NG}$ & 58.8  & 60.0  & 67.8  & 72.5  & 76.6  \\ \noalign{\smallskip}\hline\noalign{\smallskip}


Average & \textbf{59.6} & \textbf{62.0} & 66.7 & 69.2 & 74.6 \\ 
+REDA & 58.9 & 61.6 & \textbf{67.9} & \textbf{71.6} & \textbf{76.1} \\ 
+REDA$_{NG}$ & 58.9 & 60.6 & \textbf{67.9} & 71.1 & 75.6 \\ \noalign{\smallskip}\hline\noalign{\smallskip}

 \end{tabular}

\end{table*}

The average test set precision and recall, as shown in Table~\ref{tab:metrics}, may elucidate the performance gains of the augmented models over the baseline models. There are two factors at play. First, the augmented models consistently exhibit higher precision than the baseline models starting from the training size 50k. Second, the gap in recall between the baseline models and the augmented models becomes much narrower in favor of the augmented models at the same time. This suggests that the augmented models seem to learn to make significantly fewer false negatives with sufficient original training examples augmented, resulting in a sudden improvement in recall compared to the baseline models. It appears that 50k is a threshold, prior to which augmentation seems detrimental to model performance, despite the substantial increase in training examples.

\begin{table*}[h]
\centering
\caption{\label{tab:metrics}
Average test set precision and recall (\%) of the four classification models trained on LCQMC's train sets of varying size with and without augmentation. Best performance given a train set size and a metric is highlighted in \textbf{bold}.}

\begin{tabular}{p{2.5cm}p{1cm}p{1cm}p{1cm}p{1cm}p{0.6cm}}
\noalign{\smallskip}\hline\noalign{\smallskip}

Models & 5k & 10k & 50k & 100k & 240k \\ \noalign{\smallskip}\hline\noalign{\smallskip}

Precison & \textbf{57.4} & 59.2 & 62.8 & 64.3 & 69.0   \\ 
+REDA & 56.8 & \textbf{59.5} & 64.1 & \textbf{66.8} & \textbf{70.3}  \\ 
+REDA$_{NG}$ & \textbf{57.4} & 58.2 & \textbf{64.4} & 66.4 & 69.9 \\ \noalign{\smallskip}\hline\noalign{\smallskip}

Recall & \textbf{75.2} & \textbf{77.5} & \textbf{82.0} & \textbf{86.2} & 89.2  \\ 
+REDA & 73.8 & 72.7 & 81.2 & 85.8 & \textbf{90.4}  \\ 
+REDA$_{NG}$ & 71.1 & 76.1 & 79.9 & 85.5 & 89.8 \\ \noalign{\smallskip}\hline\noalign{\smallskip}


 \end{tabular}

\end{table*}

 \subsection{English: QQQD}

The test set accuracy on QQQD shown in Table~\ref{tab:qqqd_accu} exhibits a similar pattern to that on LCQMC for two reasons. First, the difference between REDA and REDA$_{NG}$ models remains negligible, reaffirming the trivial role probabilistic linguistic knowledge plays in the five token-level text augmentation techniques. Second, the augmented models do not outperform the baseline models until a sufficient number of original training examples are seen. However, unlike the experiment on LCQMC, this time the REDA$_{NG}$ models consistently perform better than the REDA models. Moreover, it appears that the threshold for the REDA and REDA$_{NG}$ models to outperform the baseline models is much larger, or 100k and 150k, respectively. These two differences are likely attributable to some training artifacts related to the datasets, the pretrained $n$-gram language models, and the likes. 

Also differing from the LCQMC experiment is how the baseline models compare to the augmented models in terms of average test set precision and recall. Instead of displaying a consistent advantage over the augmented models in one of these two metrics, the baseline models show better precision in a way highly correlated with accuracy. In other words, the augmented models do not outperform the baseline models until their respective thresholds. This indicates that the baseline models tend to make fewer false positives, compared to the augmented models when the original training data is insufficient for the text augmentation to become effective.

\begin{table*}[h]
\centering
\caption{\label{tab:qqqd_accu}
Test set accuracy (\%) of the four classification models trained on QQQD's train sets of varying size with and without augmentation.}

\begin{tabular}{p{2.5cm}p{1cm}p{1cm}p{1cm}p{1cm}p{0.6cm}}
\noalign{\smallskip}\hline\noalign{\smallskip}
Models & 10k & 50k & 100k & 150k & 260k \\ 
\noalign{\smallskip}\hline\noalign{\smallskip}

CBOW & \textbf{64.4}  & \textbf{69.9}  & 72.1  & 74.2  & 77.7 \\ 
+REDA & 62.5  & 68.5  & 71.6  & 74.8  & 78.0  \\ 
+REDA$_{NG}$ & 62.9  & 69.4  & \textbf{74.0}  & \textbf{75.5}  & \textbf{78.2}  \\ 
\noalign{\smallskip}\hline\noalign{\smallskip}

CNN & \textbf{66.1}  & \textbf{71.1}  & 72.6  & 73.4  & 75.9  \\ 
+REDA & 63.7  & 69.9  & \textbf{72.7}  & \textbf{75.3}  & 77.6  \\ 
+REDA$_{NG}$ & 63.5  & 69.3  & \textbf{72.7}  & 74.7  & \textbf{77.7} \\ 
\noalign{\smallskip}\hline\noalign{\smallskip}

LSTM & \textbf{65.7}  & \textbf{71.6}  & \textbf{72.9}  & 75.0  & 77.9 \\ 
+REDA & 64.0  & 69.8  & 72.5  & \textbf{75.1}  & \textbf{78.1}  \\ 
+REDA$_{NG}$ & 64.9  & 70.3  & 72.7  & 75.0  & \textbf{78.1} \\ 
\noalign{\smallskip}\hline\noalign{\smallskip}

GRU & \textbf{67.2}  & \textbf{71.0}  & \textbf{74.3}  & 74.7  & 77.4 \\ 
+REDA & 63.3  & 70.0  & 72.8  & 74.8  & 78.1 \\ 
+REDA$_{NG}$ & 64.0  & 70.2  & 73.8  & \textbf{75.7}  & \textbf{78.9} \\ 
\noalign{\smallskip}\hline\noalign{\smallskip}

Average & \textbf{65.9} & \textbf{70.9}  & 73.0  & 74.3  & 77.2 \\ 
+REDA & 63.4  & 69.6  & 72.4  & 75.0  & 78.0 \\ 
+REDA$_{NG}$ & 63.8  & 69.8  & \textbf{73.3}  & \textbf{75.2}  & \textbf{78.2} \\ 
\noalign{\smallskip}\hline\noalign{\smallskip}
 \end{tabular}
\end{table*}










\begin{table*}[h]
\centering
\caption{\label{tab:qqqd_metrics}
Average test set precision and recall (\%) of the four classification models trained on QQQD's train sets of varying size with and without augmentation.}

\begin{tabular}{p{2.5cm}p{1cm}p{1cm}p{1cm}p{1cm}p{0.6cm}}
\noalign{\smallskip}\hline\noalign{\smallskip}

Models & 10k & 50k & 100k & 150k & 260k \\ \noalign{\smallskip}\hline\noalign{\smallskip}

Precison & \textbf{65.0} & \textbf{69.8} & 71.6 & 72.1 & 76.0   \\ 
+REDA & 61.8 & 68.0 & 70.6 & 73.6 & 76.2  \\ 
+REDA$_{NG}$ & 62.6 & 69.0 & \textbf{72.3} & \textbf{74.2} & \textbf{77.3} \\ \noalign{\smallskip}\hline\noalign{\smallskip}

Recall & 69.5 & 73.6 & 76.2 & \textbf{79.4} & 79.6  \\ 
+REDA & \textbf{70.4} & \textbf{73.7} & \textbf{77.0} & 78.0 & \textbf{81.4}  \\ 
+REDA$_{NG}$ & 69.2 & 72.0 & 75.7 & 77.4 & 80.0 \\ \noalign{\smallskip}\hline\noalign{\smallskip}


 \end{tabular}

\end{table*}

\subsection{Interim summary}

Overall, the results presented above demonstrate that incorporating probabilistic linguistic knowledge into REDA does not make a significant difference. Pairwise Mann-Whitney U tests confirm that there is no statistically significant difference in the test set performance between the REDA and REDA$_{NG}$ models, with the obtained $p$-values close to 1.0, regardless of the specific metric in use.

Additionally, it is revealed that the five token-level text augmentation techniques are not always effective, irrespective of whether an $n$-gram language model is employed to optimize the augmented outputs or not. The results indicate that for both Chinese and English binary question matching tasks, the augmented models only outperform the baseline models when a sufficient amount of original training examples are augmented.

There are two differences observed between the experiments in Chinese and English. The differences concern the relative performance of the REDA$_{NG}$ models against the REDA models, and the relative performance of the augmented models against the baseline models. Nevertheless, these differences do not impact the two general observations made above for the purpose of this study.

\section{Supplementary experiments\label{sec:sup_exps}}

Following the main results obtained in Sec~\ref{sec:main_results}, three important follow-up questions arise: (1) Does REDA$_{NG}$ truly produce higher-quality augmented texts than REDA under the same conditions? (2) Would the results remain valid if state-of-the-art transformer models were employed instead? (3) What if the five token-level text augmentation techniques were applied separately, rather than together? Question (1) is crucial because it determines whether the insignificant difference between the REDA and REDA$_{NG}$ models found in Sec~\ref{sec:main_results} is due to the marginal role of probabilistic linguistic knowledge, or  simply because the texts augmented by REDA and REDA$_{NG}$ are indistinguishable in terms of quality. Questions (2) and (3) assess the generality of the observations made so far.   

Due to resource constraints and for simplicity, the supplementary experiments in this section are based on LCQMC.

\subsection{Comparison of texts augmented by REDA and REDA$_{NG}$ \label{sec:text_quality}}

Directly comparing the texts augmented by REDA and REDA$_{NG}$ is not feasible, three text restoration experiments were therefore designed to approximate the comparison. These experiments assess the ability of both programs to restore natural texts when given distorted texts or a pseudo-synonym dictionary for the following text editing operations: Synonym Replacement (SR), Random Swap (RS), and Random Deletion (RD). Random Insertion (RI) and Random Mix (RM) are omitted since inserting random synonyms is generally not representative of natural language use, and the text quality resulting from RM can be inferred from the other basic operations.

\begin{table}
\centering

\caption{\label{tab:quality_experiment}
Average accuracy (\%) in three text restoration tasks based on different number of edits (header). SR: Synonym Replacement; RS: Random Swap; RD: Random Deletion. Best performance given an edit number and an augmentation method is highlighted in \textbf{bold}.}
\begin{tabular}{p{1cm}p{2cm}p{1cm}p{1cm}p{0.7cm}}
\noalign{\smallskip}\hline\noalign{\smallskip}

&  & One & Two & Three \\ \noalign{\smallskip}\hline\noalign{\smallskip}
SR & REDA & 22  & 6  & 2  \\ 
& REDA$_{NG}$ & \textbf{88 } & \textbf{79}  & \textbf{64}  \\ \noalign{\smallskip}\hline\noalign{\smallskip}
RS & REDA & 9  & 4  & 4  \\ 
& REDA$_{NG}$ & \textbf{69 } & \textbf{41}  & \textbf{34}  \\ \noalign{\smallskip}\hline\noalign{\smallskip}
RD & REDA & 16  & 5  & 2  \\ 
& REDA$_{NG}$ & \textbf{39} & \textbf{22}  & \textbf{15}  \\ \noalign{\smallskip}\hline\noalign{\smallskip}
 \end{tabular}

\end{table}

Table~\ref{tab:quality_experiment} presents the average accuracy, with the experiment details provided in Appendix A. As shown, while the performance for both approaches declines as the number of edits increases, REDA$_{NG}$ consistently outperforms REDA. For REDA, restoring the distorted texts to their original form is merely a matter of chance, equal to the reciprocal of the number of possible augmented outputs. However, REDA$_{NG}$ augments texts based on the maximum likelihood principle, which tends to be closer to natural texts. This also holds true when natural texts are used as inputs. For instance, through manual inspection, I found that REDA$_{NG}$ performed much better in selecting appropriate synonyms, a problem for REDA due to its randomness and the ubiquitous existence of polysemy. By measuring the bigram overlap rate and Levenshtein edit distances of output texts randomly swapped twice from the natural texts, I further found that the average overlap rate for REDA was much lower (i.e., 0.29 versus 0.77), and that the average edit distances were much larger (i.e., 3.0 versus 1.4) than REDA$_{NG}$. This suggests that REDA$_{NG}$ preserves more collocational features of natural texts than REDA and thus augments higher-quality texts. 

 \subsection{Effect of transformer}

ERNIE-Gram \cite{xiao2020ernie-gram} is a transformer-based pretrained large language model and was chosen for its state-of-the-art performance on LCQMC. The fine-tuning of the ERNIE-Gram models shared identical training details with the main experiments, except that a smaller learning rate (i.e., 5e-5) was used. Table~\ref{tab:tranformer_results} shows the test set performance across the three metrics for the fine-tuned ERNIE-Gram models. 






\begin{table*}[h]
\centering
\caption{\label{tab:tranformer_results}
Test set, accuracy, precision and recall (\%) for the Ernie-Gram models fine-tuned on LCQMC's train sets of varying size with and without augmentation.}

\begin{tabular}{p{2.5cm}p{1cm}p{1cm}p{1cm}p{1cm}p{0.6cm}}
\noalign{\smallskip}\hline\noalign{\smallskip}

Models & 5k & 10k & 50k & 100k & 240k \\ \noalign{\smallskip}\hline\noalign{\smallskip}

Accuracy & \textbf{78.7} & \textbf{81.7} & \textbf{85.9} & \textbf{87.1} & \textbf{87.4} \\ 
+REDA & 77.5  & 80.3  & 84.1  & 85.0  & 85.7 \\ 
+REDA$_{NG}$ & 78.6  & 80.1  & 83.8  & 84.6  & 85.8 \\ 
\noalign{\smallskip}\hline\noalign{\smallskip}

Precison & \textbf{71.2} & \textbf{74.7} & \textbf{80.3} & \textbf{81.7} & \textbf{82.0}  \\ 
+REDA & 70.0 & 72.9 & 77.7 & 79.1 & 79.5  \\ 
+REDA$_{NG}$ & 71.0 & 73.2 & 77.4 & 78.6 & 80.0  \\ \noalign{\smallskip}\hline\noalign{\smallskip}

Recall & \textbf{96.6} & 95.9 & 95.2 & \textbf{95.6} & 95.9  \\ 
+REDA & 96.4 & \textbf{96.4} & \textbf{95.6} & 95.1 & \textbf{96.1}  \\ 
+REDA$_{NG}$ & 95.7 & 95.1 & 95.3 & 95.1 & 95.5 \\ \noalign{\smallskip}\hline\noalign{\smallskip}


 \end{tabular}

\end{table*}

Not surprisingly, the fine-tuned ERNIE-Gram models achieve significantly better results than the four classification models trained in the main experiments on LCQMC. Notably, using only 5k original examples, the fine-tuned ERNIE-Gram models outperform any model trained in the main experiments, regardless of augmentation. This highlights the impressive effectiveness of transfer learning resulting from fine-tuning large language model on downstream tasks. The implication may be that transfer learning is a more robust and effective way of boosting model performance than the text augmentation approaches considered in this study. However, it remains unknown whether this is also the case in low-resource settings. 

Despite the noticeable performance gain, the ERNIE-Gram models fine-tuned on augmented train sets are consistently outperformed by the baseline models without augmentation in terms of both accuracy and precision in the test set. Thus, both text augmentation approaches appear to be overall detrimental to model performance. Furthermore, no evidence indicates a significant difference between REDA and REDA$_{NG}$, even when a transformer, such as ERNIE-Gram, is used.

 \subsection{Effect of single augmentation technique}

To understand the role of each text augmentation technique, models were trained on train sets augmented using only one augmentation technique. The original train set was partitioned into 11 different sizes, rather than 5, to validate the observation in Sec~\ref{sec:main_results} that the effectiveness of the augmentation is restricted to a sufficient number of original training examples. The experimental details can be found in Appendix B.

Fig~\ref{fig:abl_accu} displays the average test set accuracy of the four classification models trained on the three types of train sets under different text augmentation techniques and across various training sizes. In line with the previous findings, the effect of probabilistic linguistic knowledge on each of the five techniques is minimal and shows no statistically significant difference, both individually and on average. Also consistent with the previous findings is the existence of a threshold where the augmented models outperform the baseline models in test set accuracy, which appears to be around the training size 100k, rather than 50k as in the related main experiments. The discrepancy may be explained by the different epoch numbers (see Appendix B) and, more importantly, the separation of the augmentation techniques, which, however, are beyond the scope of this study.

\begin{center}

\begin{figure*}[h]
\sidecaption
\includegraphics[scale=.5]{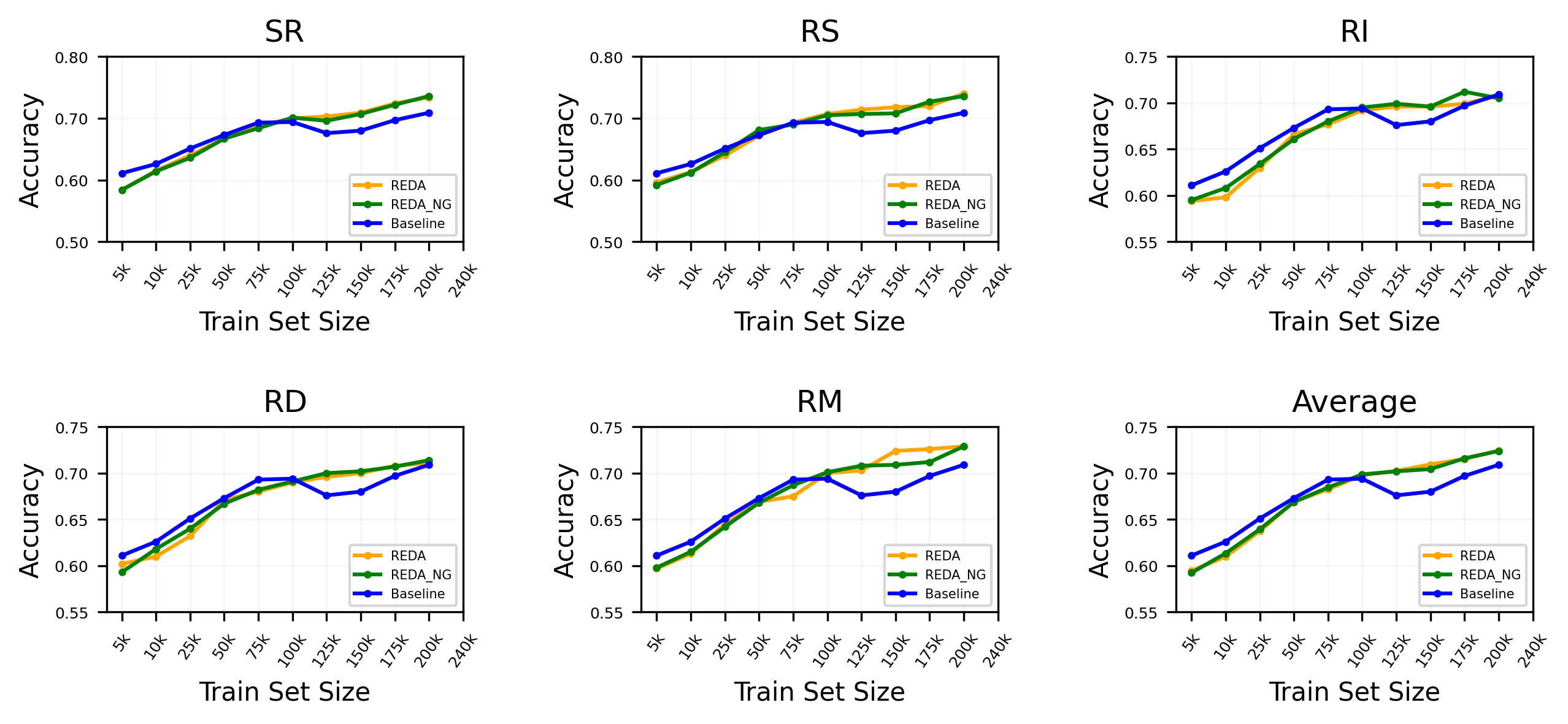}
%
%
\caption{Average test set accuracy of the four classification models trained on LCQMC’s train sets of varying size with and without augmentation under different conditions (i.e., augmentation technique, train size size). The sixth plot averages the statistics of the previous five plots.}
\label{fig:abl_accu}      
\end{figure*}
    
\end{center}

The average test set precision and recall resemble the data patterns observed in the main experiments, with an updated threshold mentioned above. Please refer to Appendix B for details.

\section{Discussion and conclusion\label{sec:discussion_conclusion}}

In this study, I evaluate the effectiveness of five token-level text augmentation techniques and the role of probabilistic linguistic knowledge thereof. To this end, two related programs were created: REDA and REDA$_{NG}$, the latter of which utilizes pretrained $n$-gram language models to select most likely augmented texts from REDA's output. Experiments on binary question matching classification task in Chinese and English strongly indicate that the role of probabilistic linguistic knowledge for token-level text augmentation is minimal and that the related augmentation techniques are not generally effective. These two findings are further discussed as follows.

First, the difference between the REDA models and the REDA$_{NG}$ models is trivial. However, the supplementary experiment on three pseudo text restoration tasks in Sec~\ref{sec:text_quality} shows that, REDA$_{NG}$ arguably generates higher-quality augmented texts compared to REDA, as it preserves more collocational features of natural texts. An intuitively plausible explanation for the insignificant role of probabilistic linguistic knowledge may be due to the inherent inability of the five augmentation techniques to produce strictly paraphrastic augmented texts. In other words, the texts augmented by REDA and REDA$_{NG}$ are to a considerable extent comparable in the sense that they are mostly not the paraphrases of the original texts being augmented. Although the REDA$_{NG}$ models appear to be slightly better than the REDA models in the English experiments, the opposite is true for the Chinese experiments. The observed differences are highly likely to result from training artifacts. Nevertheless, none of them are statistically significant.

Second, the effectiveness of the augmentation techniques, whether applied together or separately, and irrespective of the classification model type (including transformers), only surfaces when a sufficiently large amount of original training examples are supplied. This finding shows that the effectiveness is task specific and not always positive, contrasting with \cite{wei-zou-2019-eda} and aligning with \cite{empirical_survey, longpre-etal-2020-effective}. Unlike the one-text-one-label classification tasks experimented in \cite{wei-zou-2019-eda}, question matching involves classifying a given question pair into a label to indicate the intentional similarity of the pair. As such, the task is inherently more sensitive to the semantic changes caused by text augmentation and thus arguably represents a more reliable evaluative task. The performance decline in the augmented models in cases with insufficient original training examples may be due to the negative effects of the false matching augmented text pairs generated by REDA and REDA$_{NG}$. However, with enough original training examples seen, the augmented models learn to mediate these negative effects and turn them somewhat into regularization, which helps the models generalize better. Nevertheless, the requirement of a sufficiently large number of training examples makes token-level text augmentation investigated here a less practical and preferable approach for tasks of similar nature to question matching.

One might argue that the differences between REDA/REDA$_{NG}$ and EDA \cite{wei-zou-2019-eda}, as described in Sec~\ref{sec:aug_methods}, could be a possible cause for the failure of text augmentation on small train sets in this study. Specifically, by disallowing deduplicates, REDA and REDA$_{NG}$ are more likely to produce more diverse yet non-paraphrastic augmented texts than EDA, given comparably small editing rates. This might exacerbate the negative effects of random text perturbations, thereby requiring more original training examples to mitigate such effects. However, I argue that the differences between REDA/REDA$_{NG}$ and EDA are not crucial. Since the augmented models in this study do not necessarily outperform the baseline models even with a non-trivial amount of original training examples (i.e., at least 50k) and when the number of augmentation is only 1 per text, there is no reason to believe that the augmented would perform better with fewer original training examples while having the same proportion of augmented examples. Furthermore, it is not surprising that EDA works for simple one-text-one-label classification tasks, despite producing imperfect augmented texts. The reason is exactly task specific. For example, in sentence-level sentiment analysis, the sentiment of a sentence is often captured by only few keywords \cite{Bing2012}. It follows, as long as an augmented text retains these few keywords or similar replacements, it still reasonably preserves the sentiment label of the original text even if it is grammatically problematic. The key lesson here is that token-level text augmentation may easily introduce noise to the training examples for those simple classification tasks while not causing label changes. As a result, the trained models generalize better.

Systematically and fairly evaluating a text augmentation is uneasy or even unknown. The limitations of this study are obvious, since it fails to experiment with different initializations of REDA/REDA$_{NG}$ or different configurations of the classification models, confined by available computing resources. Nevertheless, this study showcases a linguistically-motivated way of evaluating text augmentation and highlights the benefits and insights it provides. The main takeaway is that although token-level text augmentation is simple and potentially useful, it should be used with caution, particularly for complex tasks.

\begin{acknowledgement}
The paper is based on two of my previous publications \cite{wang-2022-linguistic, wang-2022-random}. I thank anonymous reviewers from ICNLSP 2022, ACL 2022 Workshop on Insights from Negative Results in NLP, and AACL-IJCNLP 2022 Workshop on Evaluation \& Comparison of NLP Systems, for their feedback. Any remaining errors are solely my responsibility. 
\end{acknowledgement}

\section*{Appendix}

\subsection*{A. Text restoration experiments \label{app:A}}

For the Synonym Replacement (SR) experiment, I created a pseudo synonym dictionary consisting of 3,855 one-word-four-synonym pairs. Each word was mapped to four pseudo synonyms, including the word itself and three non-synonym random words. All the words in the dictionary were those with frequencies ranking between the 1,000th and the 10,000th positions in the unigram dictionary compiled for the Chinese $n$-gram language model. For the Random Swap (RS) and Random Deletion (RD) experiments, I randomly reordered the natural texts and added random words from the texts before performing RS and RD, respectively. For each comparison made, I randomly sampled 10,000 texts from LCQMC's train set for five runs.

\subsection*{B. Ablation experiments on LCQMC}

The training conditions were the same as the main experiments, except for training time. Specifically, to save resources, the training time was reduced to 2 epochs when the train size was 50k or 100k, and to 1 epoch when the size was over 100k. Since the aim is to compare the test set performance among the baseline, REDA, and REDA$_{NG}$ models, and because larger train sizes require fewer epochs to fit the train sets, the reduction of the training time is considered reasonable. 

\begin{figure*}[h]
\sidecaption
\includegraphics[scale=.5]{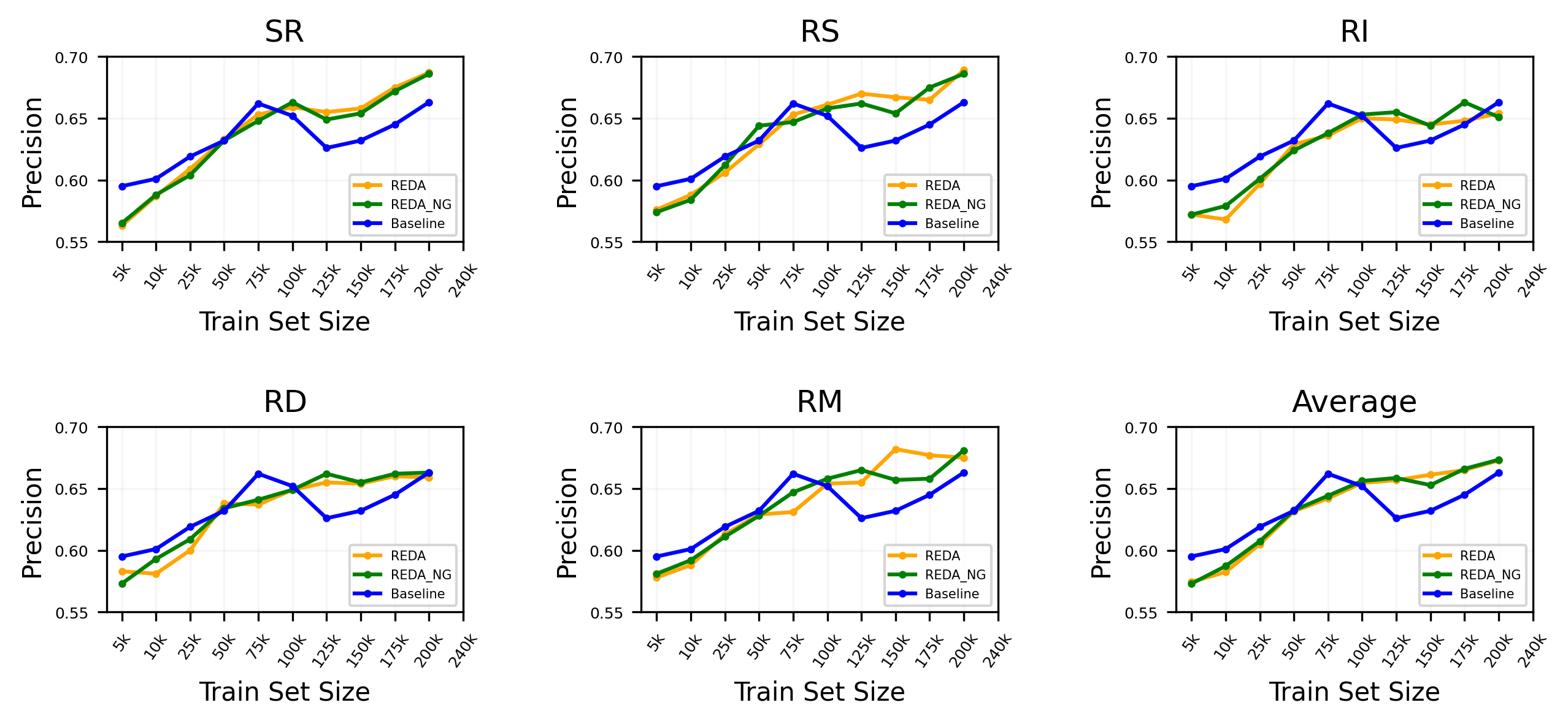}
%
%
\caption{Average test set precision of the four classification models trained on LCQMC’s train sets of varying size with and without augmentation under different conditions.}
\label{fig:abl_prec}      
\end{figure*}

\begin{figure*}[h]
\sidecaption
\includegraphics[scale=.5]{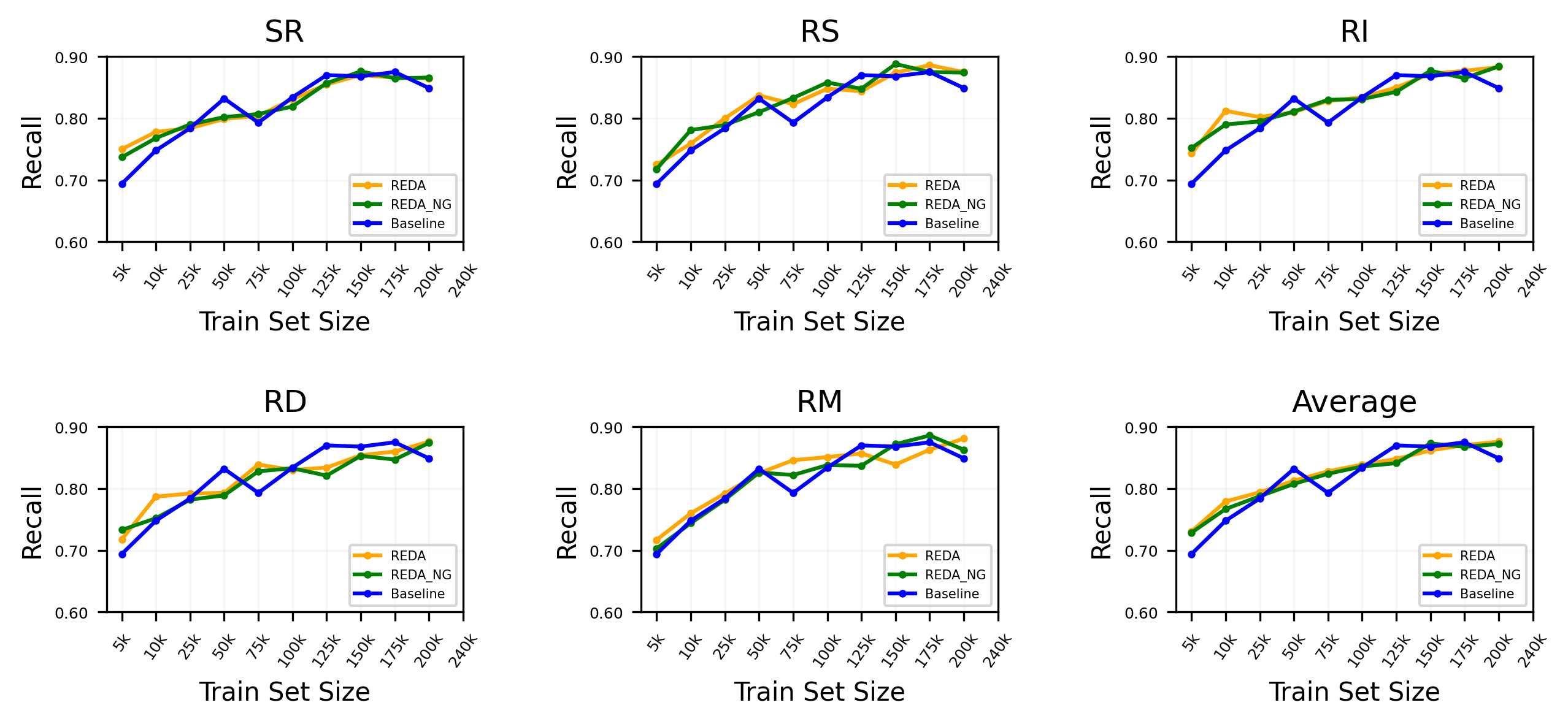}
%
%
\caption{Average test set recall of the four classification models trained on LCQMC’s train sets of varying size with and without augmentation under different conditions.}
\label{fig:abl_recall}      
\end{figure*}

For the ablation experiments, Table~\ref{tab:each_text_aug_size} displays the size of the augmented train sets, and Figs~\ref{fig:abl_prec} and~\ref{fig:abl_recall} show the average test set precision and recall, respectively.  

\begin{table}[h]
\centering
\caption{\label{tab:each_text_aug_size}
Size of augmented train sets per text augmentation technique for LCQMC.}
\begin{tabular}{p{1.5cm}p{1.5cm}p{1.5cm}p{1.5cm}p{1.5cm}p{1cm}}
\noalign{\smallskip}\hline\noalign{\smallskip}
\textbf{Size} & \textbf{SR} & \textbf{RS} & \textbf{RI} & \textbf{RD} & \textbf{RM} \\ \noalign{\smallskip}\hline\noalign{\smallskip}
5k & 24,402 & 24,758 & 16,733 & 16,780 & 24,859 \\ \noalign{\smallskip}
10k & 48,807 & 49,575 & 33,090 & 33,208 & 49,652 \\ \noalign{\smallskip}
25k & 122,358 & 124,040 & 83,329 & 83,592 & 124,237 \\ \noalign{\smallskip}
50k & 244,577 & 248,074 & 166,839 & 167,296 & 248,539 \\ \noalign{\smallskip}
75k & 220,843 & 223,497 & 162,563 & 162,972 & 224,026 \\ \noalign{\smallskip}
100k & 294,516 & 297,987 & 216,540 & 217,012 & 298,620 \\ \noalign{\smallskip}
125k & 368,078 & 372,536 & 270,957 & 271,552 & 373,266 \\ \noalign{\smallskip}
150k & 441,643 & 446,941 & 325,027 & 325,738 & 447,838 \\ \noalign{\smallskip}
175k & 515,229 & 521,484 & 379,352 & 380,214 & 522,535 \\ \noalign{\smallskip}
200k & 588,901 & 595,977 & 433,521 & 434,469 & 597,084 \\ \noalign{\smallskip}
240k & 703,077 & 711,631 & 517,492 & 518,664 & 712,852 \\ \noalign{\smallskip}\hline\noalign{\smallskip}
\end{tabular}
\end{table}

\begin{thebibliography}{99.}%

\bibitem{Simard2003}Simard, P., Steinkraus, D. \& Platt, J. Best practices for convolutional neural networks applied to visual document analysis. {\em Seventh International Conference On Document Analysis And Recognition, 2003. Proceedings.}. pp. 958-963 (2003)
\bibitem{alexnet2012}Krizhevsky, A., Sutskever, I. \& Hinton, G. ImageNet Classification with Deep Convolutional Neural Networks. {\em Advances In Neural Information Processing Systems}. \textbf{25} (2012)
\bibitem{ko15_interspeech}Ko, T., Peddinti, V., Povey, D. \& Khudanpur, S. Audio augmentation for speech recognition. {\em Proc. Interspeech 2015}. pp. 3586-3589 (2015)
\bibitem{Cui2015}Cui, X., Goel, V. \& Kingsbury, B. Data Augmentation for Deep Neural Network Acoustic Modeling. {\em IEEE/ACM Transactions On Audio, Speech, And Language Processing}. \textbf{23}, 1469-1477 (2015)
\bibitem{Park2019}Park, D., Chan, W., Zhang, Y., Chiu, C., Zoph, B., Cubuk, E. \& Le, Q. SpecAugment: A Simple Data Augmentation Method for Automatic Speech Recognition. {\em Proc. Interspeech 2019}. pp. 2613-2617 (2019)
\bibitem{Shorten2019ASO}Shorten, C. \& Khoshgoftaar, T. A survey on Image Data Augmentation for Deep Learning. {\em Journal Of Big Data}. \textbf{6} pp. 1-48 (2019)
\bibitem{Iwana2021}Iwana, S. An empirical survey of data augmentation for time series classification with neural networks. {\em PLOS ONE}. \textbf{16}, 1-32 (2021,7), https://doi.org/10.1371/journal.pone.0254841
\bibitem{Shorten2021}Shorten, C., Khoshgoftaar, T. \& Furht, B. Text Data Augmentation for Deep Learning. {\em Journal Of Big Data}. \textbf{8} pp. 1-34 (2021)
\bibitem{feng-etal-2021-survey}Feng, S., Gangal, V., Wei, J., Chandar, S., Vosoughi, S., Mitamura, T. \& Hovy, E. A Survey of Data Augmentation Approaches for NLP. {\em Findings Of The Association For Computational Linguistics: ACL-IJCNLP 2021}. pp. 968-988 (2021,8), https://aclanthology.org/2021.findings-acl.84
\bibitem{Liu:9240734}Liu, P., Wang, X., Xiang, C. \& Meng, W. A Survey of Text Data Augmentation. {\em 2020 International Conference On Computer Communication And Network Security (CCNS)}. pp. 191-195 (2020)
\bibitem{yang-etal-2022-learning}Yang, D., Parikh, A. \& Raffel, C. Learning with Limited Text Data. {\em Proceedings Of The 60th Annual Meeting Of The Association For Computational Linguistics: Tutorial Abstracts}. pp. 28-31 (2022,5), https://aclanthology.org/2022.acl-tutorials.5
\bibitem{sahin-2022-augment}Şahin, G. To Augment or Not to Augment? A Comparative Study on Text Augmentation Techniques for Low-Resource NLP. {\em Computational Linguistics}. \textbf{48}, 5-42 (2022,3), https://aclanthology.org/2022.cl-1.2
\bibitem{empirical_survey}Chen, J., Tam, D., Raffel, C., Bansal, M. \& Yang, D. An Empirical Survey of Data Augmentation for Limited Data Learning in NLP. {\em Transactions Of The Association For Computational Linguistics}. \textbf{11} pp. 191-211 (2023,3), https://doi.org/10.1162/tacl
\bibitem{Zhang2015}Zhang, X., Zhao, J. \& LeCun, Y. Character-level Convolutional Networks for Text Classification. {\em Advances In Neural Information Processing Systems}. \textbf{28} (2015), https://proceedings.neurips.cc/paper/2015/file/250cf8b51c773f3f8dc8b4be867a9a02-Paper.pdf
\bibitem{wang-yang-2015-thats}Wang, W. \& Yang, D. That's So Annoying!!!: A Lexical and Frame-Semantic Embedding Based Data Augmentation Approach to Automatic Categorization of Annoying Behaviors using \#petpeeve Tweets. {\em Proceedings Of The 2015 Conference On Empirical Methods In Natural Language Processing}. pp. 2557-2563 (2015,9), https://aclanthology.org/D15-1306
\bibitem{wei-zou-2019-eda}Wei, J. \& Zou, K. EDA: Easy Data Augmentation Techniques for Boosting Performance on Text Classification Tasks. {\em Proceedings Of The 2019 Conference On Empirical Methods In Natural Language Processing And The 9th International Joint Conference On Natural Language Processing (EMNLP-IJCNLP)}. pp. 6382-6388 (2019,11), https://aclanthology.org/D19-1670
\bibitem{kang-etal-2018-adventure}Kang, D., Khot, T., Sabharwal, A. \& Hovy, E. AdvEntuRe: Adversarial Training for Textual Entailment with Knowledge-Guided Examples. {\em Proceedings Of The 56th Annual Meeting Of The Association For Computational Linguistics (Volume 1: Long Papers)}. pp. 2418-2428 (2018,7), https://aclanthology.org/P18-1225
\bibitem{asai-hajishirzi-2020-logic}Asai, A. \& Hajishirzi, H. Logic-Guided Data Augmentation and Regularization for Consistent Question Answering. {\em Proceedings Of The 58th Annual Meeting Of The Association For Computational Linguistics}. pp. 5642-5650 (2020,7), https://aclanthology.org/2020.acl-main.499
\bibitem{sennrich-etal-2016-improving}Sennrich, R., Haddow, B. \& Birch, A. Improving Neural Machine Translation Models with Monolingual Data. {\em Proceedings Of The 54th Annual Meeting Of The Association For Computational Linguistics (Volume 1: Long Papers)}. pp. 86-96 (2016,8), https://aclanthology.org/P16-1009
\bibitem{edunov-etal-2018-understanding}Edunov, S., Ott, M., Auli, M. \& Grangier, D. Understanding Back-Translation at Scale. {\em Proceedings Of The 2018 Conference On Empirical Methods In Natural Language Processing}. pp. 489-500 (2018), https://aclanthology.org/D18-1045
\bibitem{Singh2019}Singh, J., McCann, B., Keskar, N., Xiong, C. \& Socher, R. XLDA: Cross-Lingual Data Augmentation for Natural Language Inference and Question Answering. {\em CoRR}. \textbf{abs/1905.11471} (2019), http://arxiv.org/abs/1905.11471
\bibitem{hou-etal-2018-sequence}Hou, Y., Liu, Y., Che, W. \& Liu, T. Sequence-to-Sequence Data Augmentation for Dialogue Language Understanding. {\em Proceedings Of The 27th International Conference On Computational Linguistics}. pp. 1234-1245 (2018,8), https://aclanthology.org/C18-1105
\bibitem{kobayashi-2018-contextual}Kobayashi, S. Contextual Augmentation: Data Augmentation by Words with Paradigmatic Relations. {\em Proceedings Of The 2018 Conference Of The North American Chapter Of The Association For Computational Linguistics: Human Language Technologies, Volume 2 (Short Papers)}. pp. 452-457 (2018,6), https://aclanthology.org/N18-2072
\bibitem{kurata16b_interspeech}Kurata, G., Xiang, B. \& Zhou, B. Labeled Data Generation with Encoder-Decoder LSTM for Semantic Slot Filling. {\em Proc. Interspeech 2016}. pp. 725-729 (2016)
\bibitem{chen-etal-2020-mixtext}Chen, J., Yang, Z. \& Yang, D. MixText: Linguistically-Informed Interpolation of Hidden Space for Semi-Supervised Text Classification. {\em Proceedings Of The 58th Annual Meeting Of The Association For Computational Linguistics}. pp. 2147-2157 (2020,7), https://aclanthology.org/2020.acl-main.194
\bibitem{puzzleMix2020}Kim, J., Choo, W. \& Song, H. Puzzle Mix: Exploiting Saliency and Local Statistics for Optimal Mixup. {\em Proceedings Of The 37th International Conference On Machine Learning}. (2020)
\bibitem{chen-etal-2022-doublemix}Chen, H., Han, W., Yang, D. \& Poria, S. DoubleMix: Simple Interpolation-Based Data Augmentation for Text Classification. {\em Proceedings Of The 29th International Conference On Computational Linguistics}. pp. 4622-4632 (2022,10), https://aclanthology.org/2022.coling-1.409
\bibitem{xie2020unsupervised}Xie, Q., Dai, Z., Hovy, E., Luong, T. \& Le, Q. Unsupervised Data Augmentation for Consistency Training. {\em Advances In Neural Information Processing Systems}. \textbf{33} pp. 6256-6268 (2020), https://proceedings.neurips.cc/paper/2020/file/44feb0096faa8326192570788b38c1d1-Paper.pdf
\bibitem{longpre-etal-2020-effective}Longpre, S., Wang, Y. \& DuBois, C. How Effective is Task-Agnostic Data Augmentation for Pretrained Transformers?. {\em Findings Of The Association For Computational Linguistics: EMNLP 2020}. pp. 4401-4411 (2020,11), https://aclanthology.org/2020.findings-emnlp.394
\bibitem{wang-2022-linguistic}Wang, Z. Linguistic Knowledge in Data Augmentation for Natural Language Processing: An Example on Chinese Question Matching. {\em Proceedings Of The 5th International Conference On Natural Language And Speech Processing (ICNLSP 2022)}. pp. 40-49 (2022,12), https://aclanthology.org/2022.icnlsp-1.5
\bibitem{wang-2022-random}Wang, Z. Random Text Perturbations Work, but not Always. {\em Proceedings Of The 3rd Workshop On Evaluation And Comparison Of NLP Systems}. pp. 51-57 (2022,11), https://aclanthology.org/2022.eval4nlp-1.6
\bibitem{wordnet}Miller, G. WordNet: A Lexical Database for English. {\em Commun. ACM}. \textbf{38}, 39-41 (1995,11), https://doi.org/10.1145/219717.219748
\bibitem{Jurafsky2009}Jurafsky, D. \& Martin, J. Speech and Language Processing: An Introduction to Natural Language Processing, Computational Linguistics, and Speech Recognition. (Prentice Hall PTR,2009)
\bibitem{brants-etal-2007-large}Brants, T., Popat, A., Xu, P., Och, F. \& Dean, J. Large Language Models in Machine Translation. {\em Proceedings Of The 2007 Joint Conference On Empirical Methods In Natural Language Processing And Computational Natural Language Learning (EMNLP-CoNLL)}. pp. 858-867 (2007,6), https://aclanthology.org/D07-1090
\bibitem{liu-etal-2018-lcqmc}Liu, X., Chen, Q., Deng, C., Zeng, H., Chen, J., Li, D. \& Tang, B. LCQMC:A Large-scale Chinese Question Matching Corpus. {\em Proceedings Of The 27th International Conference On Computational Linguistics}. pp. 1952-1962 (2018,8), https://aclanthology.org/C18-1166
\bibitem{cbow}Mikolov, T., Chen, K., Corrado, G. \& Dean, J. Efficient Estimation of Word Representations in Vector Space. {\em 1st International Conference On Learning Representations, ICLR 2013, Scottsdale, Arizona, USA, May 2-4, 2013, Workshop Track Proceedings}. (2013), http://arxiv.org/abs/1301.3781
\bibitem{kim-2014-convolutional}Kim, Y. Convolutional Neural Networks for Sentence Classification. {\em Proceedings Of The 2014 Conference On Empirical Methods In Natural Language Processing (EMNLP)}. pp. 1746-1751 (2014,10), https://aclanthology.org/D14-1181
\bibitem{cho-etal-2014-learning}Cho, K., Merriënboer, B., Gulcehre, C., Bahdanau, D., Bougares, F., Schwenk, H. \& Bengio, Y. Learning Phrase Representations using RNN Encoder–Decoder for Statistical Machine Translation. {\em Proceedings Of The 2014 Conference On Empirical Methods In Natural Language Processing (EMNLP)}. pp. 1724-1734 (2014,10), https://aclanthology.org/D14-1179
\bibitem{lstm}Hochreiter, S. \& Schmidhuber, J. Long Short-Term Memory. {\em Neural Comput.}. \textbf{9}, 1735-1780 (1997,11), https://doi.org/10.1162/neco.1997.9.8.1735
\bibitem{tranformer}Vaswani, A., Shazeer, N., Parmar, N., Uszkoreit, J., Jones, L., Gomez, A., Kaiser, Ł. \& Polosukhin, I. Attention is All you Need. {\em Advances In Neural Information Processing Systems}. \textbf{30} (2017), https://proceedings.neurips.cc/paper/2017/file/3f5ee243547dee91fbd053c1c4a845aa-Paper.pdf
\bibitem{adam}Kingma, D. \& Ba, J. Adam: A Method for Stochastic Optimization. {\em 3rd International Conference On Learning Representations, ICLR 2015, San Diego, CA, USA, May 7-9, 2015, Conference Track Proceedings}. (2015), http://arxiv.org/abs/1412.6980
\bibitem{xiao2020ernie-gram}Xiao, D., Li, Y., Zhang, H., Sun, Y., Tian, H., Wu, H. \& Wang, H. ERNIE-Gram: Pre-Training with Explicitly N-Gram Masked Language Modeling for Natural Language Understanding. {\em Proceedings Of The 2021 Conference Of The North American Chapter Of The Association For Computational Linguistics: Human Language Technologies}. pp. 1702-1715 (2021,6), https://aclanthology.org/2021.naacl-main.136
\bibitem{Bing2012}Liu, B. Sentiment Analysis and Opinion Mining. (Morgan \& Claypool,2012)


\end{thebibliography}
\end{document}